# A GEOMETRICALLY CONSTRAINED POINT MATCHING BASED ON VIEW-INVARIANT CROSS-RATIOS, AND HOMOGRAPHY

*Yueh-Cheng Huang, Ching-Huai Yang, Chen-Tao Hsu, and Jen-Hui Chuang*
*Dept. of Computer Science, National Yang Ming Chiao Tung Univ., Hsinchu, Taiwan*

**ABSTRACT**

In computer vision, finding point correspondence among images plays an important role in many applications, such as image stitching, image retrieval, visual localization, etc. Most of the research works focus on the matching of local feature before a sampling method is employed, such as RANSAC, to verify initial matching results via repeated fitting of certain global transformation among the images. However, incorrect matches may still exist, while careful examination of such problems is often skipped. Accordingly, a geometrically constrained algorithm is proposed in this work to verify the correctness of initially matched SIFT keypoints based on view-invariant cross-ratios (CRs). By randomly forming pentagons from these keypoints and matching their shape and location among images with CRs, robust planar region estimation can be achieved efficiently for the above verification, while correct and incorrect matches of keypoints can be examined easily with respect to those shape and location matched pentagons. Experimental results show that satisfactory results can be obtained for various scenes with single as well as multiple planar regions.

*Index Terms:* point correspondence, coplanar regions, outlier removal, view-invariant cross-ratio

## 1. INTRODUCTION

In computer vision, identification of feature points correspondence between two images plays an important role in many applications, such as image stitching, image retrieval, visual localization, etc. Usually, the first step of the foregoing identification is to extract image feature points using certain descriptors of feature point. Numerous algorithms for feature descriptors, such as SIFT [1], SURF [2], ORB [3], GLOH [4], and BOLD, [5] use local image characteristics to extract the most unique and stable keypoints for images from different views, before trying to match them across multiple images. However, local feature extraction and matching methods may generate erroneous correspondence (outliers) at times.

In general, as indicated in [6], there are two main types of algorithms for correspondence verification, i.e., geometric verification methods and the learning-based methods, with performance of the latter not yet well understood. Besides, learning-based algorithms, such as outlier classification [7, 8], deep fundamental matrix estimation from data [9], or neural-guided RANSAC [10], often cause high computation cost. Thus, only geometric methods are considered here.

Geometric verification methods [11, 12, 13] identify correct point correspondences, and removing outliers, by fitting certain global transformation, e.g., similarity, affinity, homography, or with fundamental matrix, to all initial point matches with sampling methods based on RANSAC [14] and its variants. In addition, fast approximate spatial verifications are adopted [15, 16, 17] to improve the efficiency. While all these methods consider the matching of planar feature commonly seen in real scenes, the perspective projection-based homography associated with a planar surface (or the camera view) is not always included in the matching because of its complexity, not to mention the multi-plane situations.

In this paper, a new way of establishing planar constraint for point correspondence verification is proposed, which uses the view-invariant cross-ratios (CRs) associated with vertices of a pentagonal region [18], as shown in Fig. 1, in place of the homography. Main features of such approach include:

I. **Efficient**: it is simpler than homography in computation complexity by more an order of magnitude.
II. **General**: with I, it is well applicable to more complex multi-plane scenes through planar region aggregation.
III. **Robust**: very stable verification results can be obtained via view-independent thresholding in the sampling.
IV. **Observable**: correct/incorrect point correspondence can be observed easily with respect to pentagons (cf. using lines connecting matched points as in Fig. 2 (a)).

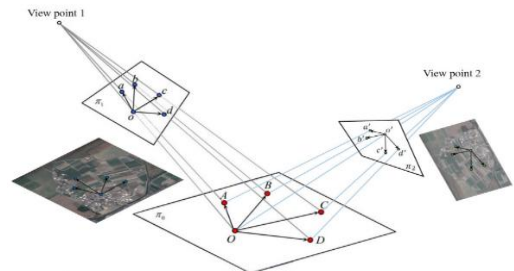

Fig. 1. Geometry for computing the view-invariant cross-ratio with respect to a vertex ($O$) of a pentagon, which can be used, together with those with respect to the other four vertices, to identify correct correspondences of pentagonal region in stereo images (see Equations (1) and (2)).

## 2. METHOD

To attain the foregoing features of point (and region) matching, major steps of the proposed verification method for point correspondences between two images ($I_1$ and $I_2$), given an initial set of point correspondences, include:

(i) Partition $I_1$ into $N \times N$ blocks. For a selected pentagonal region in each block, check for its correct shape in $I_2$.

(ii) Identify planar regions in $I_1$ (and $I_2$) by merging coplanar pentagonal regions verified in (i).

(iii) Identify other correctly matched points in $I_1$ (and $I_2$) with the planar constraints established in (ii).

Fig. 2 shows an initial set of point correspondences obtained using SIFT [1], and point/region matching results obtained with the proposed approach wherein a single planar region, depicted with (red, not blue) pentagons and matched feature points for better visualization, is identified in two images. Next, implementation of the foregoing verification procedure will be elaborated, followed by a complexity comparison between utilizing CRs and homography for the large amount of checks in (i), and the merger in (ii).

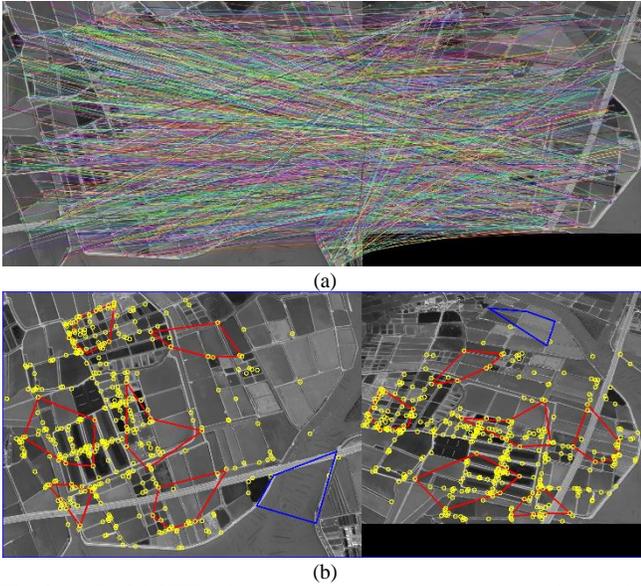

Fig. 2. (a) Initial SIFT point correspondences, (b) our easy-to-check point (and region) matching results based on a 3×3 partition of the left image.

### 2.1 View-invariant pentagonal shape matching in (i)

To work with corresponding scenes, say in $I_1$ and $I_2$, which contain multiple planar regions, one of the stereo images, say $I_1$, is first partitioned into $N \times N$ blocks wherein the shape matching process can be performed for each block for different camera views[1]. To that end, the view-invariant CR is employed in the matching for a selected pentagonal region. For example, with point $O$ being the origin of position vectors of the other four vertices in Fig. 1, the two CRs

$$CR_o(a,b,c,d) = \frac{|\vec{a} \times \vec{c}||\vec{b} \times \vec{d}|}{|\vec{b} \times \vec{c}||\vec{a} \times \vec{d}|} \quad (1)$$

and

$$CR_{o'}(a',b',c',d') = \frac{|\vec{a'} \times \vec{c'}||\vec{b'} \times \vec{d'}|}{|\vec{b'} \times \vec{c'}||\vec{a'} \times \vec{d'}|} \quad (2)$$

can be defined in View 1 and View 2, respectively. Being invariant under perspective projection, the two CRs should be equal in theory. Accordingly, such idea is applied in the following matching procedure for a pentagon formed by five selected keypoints in an image block of $I_1$:

1. Randomly select five keypoints, together with their matches in $I_2$, from the initial set of point correspondences.
2. Obtain five pairs of *CR-CR'* for all corresponding vertices of the above two pentagons, one in $I_1$ and the other in $I_2$.
3. Report a successful shape matching if each *CR-CR'* pair are *almost the same*.
4. Report an unsuccessful matching if a predetermined trial count is reached; otherwise initiate the next matching trial.

Specifically, for the example shown in Fig. 1, $CR_o$ and $CR'_{o'}$ are considered almost the same in Step 3 if

$$\frac{|CR_o - CR_{o'}|}{|CR_o + CR_{o'}|} \leq 5\% \quad (3)$$

Fig. 2 (b) shows some shape matched pentagonal regions thus obtained for a trial count of $1,000^2$. The number of planar regions in the scene will then be determined by merging the above matched pentagons into groups, each corresponds to one planar region, as discussed next. (More discussion on various trial count settings will be given in experiments.)

### 2.2 Identifying planar regions for (ii)

Major steps of merging matched pentagons obtained in the previous subsection into coplanar regions in $I_1$ and $I_2$ can be summarized in the following steps:

A. Identify pairs of mistakenly matched pentagonal regions and remove them from subsequent processes.
B. Incrementally merge the remaining pentagonal regions in $I_1$ (and $I_2$) into a planar group, also using *CR*s as before.
C. Establish additional planar groups for pentagonal regions which cannot be merged into existing planar groups.
D. End when all matched pentagon pairs are processed.

For the shape matched pentagon pairs shown in Fig. 2, one (blue) pair corresponds to erroneous matches, e.g., with unmatched size and/or location, and will be removed from the final (red) pentagon pairs by Step A. One can see from this typical case that as $I_1$ is partitioned with a 3×3 grid, erroneous matches can be detected easily based on rather simple rules of geometry (instead of considering their precise locations as in Sec. 2.3). For example, the 3×3 adjacency relationship among the pentagons in $I_2$ of Fig. 2 should be identical to that among their counterparts in $I_1$.

---

[1] The whole image can be used directly for the matching without any partition if the scene is known to have only a single planar region.

[2] While more trials may result in more matches, only a single pentagon pair is needed in theory for identifying correct point correspondences in each planar region, as discussed in Sec. 2.3.

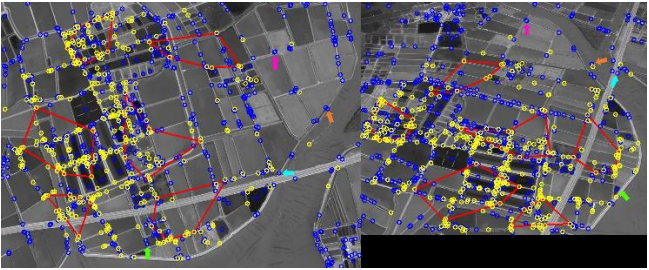

Fig. 3. Identification of correct (same as in Fig. 2) and incorrect (blue) point matches for all initially matched keypoints, with some typical cases of the latter indicated with arrows of different color.

For the example shown in Fig. 2, a single planar region[3] is then established with Step B by merging the remaining six good matches, with the final merged result depicted with a single (red) color. As a matter of fact, Step B corresponds to a shape matching procedure similar to the that presented in Sec. 2.1. In particular, it is implemented by the following procedure for two pairs of pentagon, say $P_1$ and $P_2$:

- ✧ Randomly pick $m$ ($n$), e.g., $m=3$, corresponding vertices from $P_1$ ($P_2$), with $m+n=5$, to form a new pentagon pair.
- ✧ Form a second pentagon pair with the remaining vertices from $P_1$ and $P_2$ (optional).
- ✧ Aggregate $P_1$ and $P_2$ into a coplanar group if each of the above pentagon pairs has matched shape.

While adjacent pentagons can be assembled with the above procedure to extend a planar region incrementally, merging them with an arbitrary order will work as well (as the same planar constraint will be established). Examples for which Step C needs to be performed to established multiple planar regions will be provided in the experiments.

**2.3 Verification of planar point correspondence in (iii)**

With the planar regions identified with Steps A~D in Sec. 2.2, the rest correct (and incorrect) correspondences of feature points in $I_1$ and $I_2$ can be identified with respect to the established set of planar constraints, one for each planar region. To that end, the planar coordinate transformation based on homography is employed as follows:

1. Establish homographic transformation between $I_1$ and $I_2$ for each pair of corresponding planar regions.
2. For each pair of initially matched feature points, if their positions are predicted accurately by a homography found above, i.e., within 10 pixels, attach them to corresponding planar region; otherwise, mark them as an incorrect match.

For example, Fig. 3 shows correct and incorrect matches thus obtained for initially matched keypoints in Fig. 2. While the former can be verified visually by examining their relative locations to nearby pentagon(s), the latter do not have similar property. Overall, 557 incorrect matches are identified for the initial 1150 matches found by BF-match [21] for SIFT keypoints obtained by OpenCV 4.5.5.6, showing a serious problem often overlooked by previous research works (and the effectiveness of the proposed approach in fixing it).

**2.4 Selection between cross-ratio and homography**

In Secs. 2.1 and 2.2, cross-ratios associated with a pentagon is selected in the computation for (i) efficiency and (ii) robustness. For (i), to determine the planar constraint from four points for the fifth, as in Eq. (1), 10 multiplications (including four outer products) and one division is needed, while a 9×9 matrix needs to be established with 648 ($9\times9\times2\times n, n = 4$) multiplications and 72 additions, before nine homography parameters can be found as its eigenvector with $O(n^3)$ in complexity [19], or down to $O(n^2 \sim n^{2.3})$ under certain conditions [20], for $n \geq 4$. As for (ii), a view-invariant threshold, as in Eq. (3), can be used, while setting such threshold is not straightforward for homography.

On the other hand, homography is used in Sec. 2.3 for the identification of correct and incorrect matches. This is because only one homography needs to established for each planar region (in Secs. 2.1 and 2.2 one homography need to be found for each pentagon in each image, and for each trial). For example, to check a match pair of feature points, its location in $I_2$ (and $I_1$ if needed) is compared with the one predicted by the homogrphy from $I_1$ to $I_2$ (9 multiplications) according to Step 2 of Sec. 2.3. Although similar check can be performed by computing CR twice[4], four references points and two position vectors need to be selected.

### 3. EXPERIMENT

As the efficiency of proposed approach is explained in Sec. 2.4, with some of its point matching results provided in an easy-to-check form for visual evaluation, additional examples are presented in this section to show its (I) robustness and (II) generality. While various parameters of our system will be evaluated for (I), realistic scenes having more than a single planar region will be considered for (II).

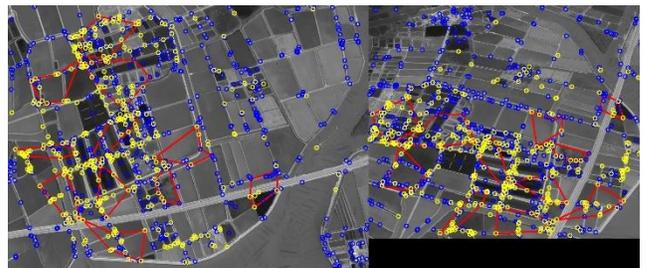

Fig. 4. Identification of correct and incorrect matches for all initially matched key points by 5×5 partition. The numbers of correct/incorrect matches are identical to those shown in Fig. 3.

For (I), a different partition, based on a 5×5 grid, is first considered for the 3×3 example shown in Fig. 3, which gives identical verification results, as shown in Fig. 4, showing the robustness of our method in establishing consistent planar constraints. Moreover, Table 1 lists "correct" (and incorrect) matches obtained with different thresholds used for (i) cross-ratio in Eq. (3) and (ii) location error in Sec. 2.3. One can see that the established planar constraint is quite consistent for variation in (i), resulting in little changes in the verification results for keypoint matching, while such results will change monotonically, as expected, with variation in (ii).

---

[3] Additional multi-plane examples will be provided in the experiments

[4] For example, one can see that the location of $o$ in view 1 of Fig. 1 can be determined by $-\vec{a}$ and $-\vec{b}$, each obtained by computing a cross-ratio.

Table 1. Correct/incorrect matches vs. different parameters.

| Allowed location error $\rightarrow$ | 8 pixels | 10 pixels | 12 pixels |
|---|---|---|---|
| $(CR-CR')/CR+CR') < 3\%$ | 586/564 | 592/558 | 597/553 |
| $(CR-CR')/CR+CR') < 5\%$ | 587/563 | 593/557 | 597/553 |
| $(CR-CR')/CR+CR') < 7\%$ | 587/563 | 592/558 | 597/553 |

As for (II), examples of more general scenes which have multiple planar regions are considered in the following. Fig. 5 shows two images of a corner wherein four shape matched pentagons are obtained for a 5×5 partition of the left image. The pentagons are then merged into two (red and orange) groups by the procedure presented in Sec. 2.2, with correctly matched keypoints identified for the two planar regions with blue and magenta circles, respectively. Overall, 285 (out of 1134) good matches are identified, with 108 identified for the ground plane and 177 for the wall. As only a small part of this complex scene is visible from both camera views, near 75% of initial matches are incorrect!

Fig. 6 shows an example for frontal views of a building, wherein a relatively large number of 666 incorrect matches are identified out of 1046 initially matched keypoints. Such results are due to less precise planar approximation of the façade of the building which has a complex (near planar) structure versus the relatively small camera-object distance, and also requires more processing time, as discussed next.

Although the trial count used in Sec. 2.1 has little effect on the verification of keypoint matching if planar constraints are established correctly[5], its value does have direct influence on computation time which is roughly proportional to amount of the execution of Eq. (3), as shown in Fig. 7 for eight tests adapted from previous examples, each executed 10 times with i7-4790K and 16GB RAM under Windows10. For each test, a trial count of 500, 1000, 1500, or 2000 is selected so that reasonable number of matched pentagons, on the average, can be obtained, with the highest computation time (up to 18 seconds) needed for images shown in Fig. 6[6].

## 4. CONCLUSION

In this paper, a globally and geometrically constrained scheme is proposed to verify the correctness of initially matched SIFT keypoints based on view-invariant CRs. To that end, robust planar region estimation is first performed by randomly forming pentagons from these keypoints and matching their shape and location among images with efficient CR evaluation. Then, homography associated with each planar region is employed to verify the correctness of every initially matched keypoint pairs. Unlike previous works, our verification results can be evaluated easily by visually examining their relative locations among matched pentagons. Moreover, the robustness of the verification with respect to different system parameters and the applicability of the approach to cope with scenes with multiple planar regions are demonstrated in the experiments. Further improvements on the efficiency, e.g., developing different ways of reducing the trial count, and the applicability of our approach to more complex scenes are currently under investigation.

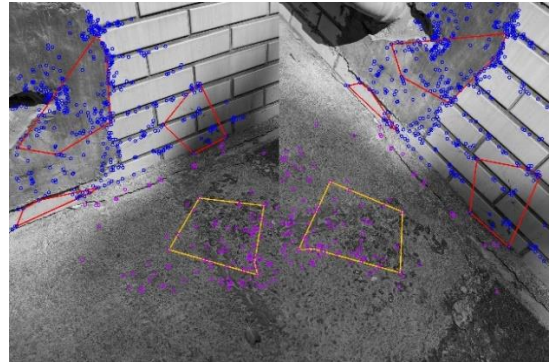

Fig. 5. Point matching results for two 1108×1477 images of a corner.

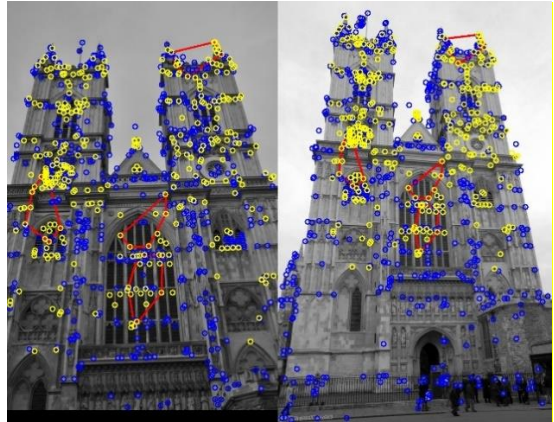

Fig. 6. Point matching results for two images of a building.

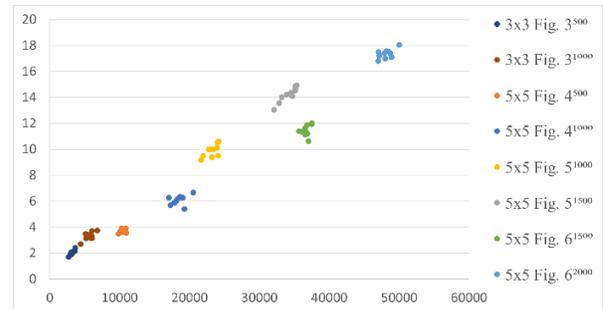

Fig. 7. Computation time (sec) vs. amount of execution of Eq. (3) for images pairs shown in Figs. 3 to 6, with superscript showing trial count setting for each group of 10 trials. The average number of correctly matched pentagon pairs for each group (from top to bottom of the depicted list): 5.0, 4.1, 6.8, 9.6, 3.0, 3.5, 1.0, 2.0.

---

[5] A trial count of 1000 is used for all examples discussed so far.

[6] Discussion on other factors which may also affect the computation time is omitted for brevity

# REFERENCES

[1] D. G. Lowe, "Distinctive Image Features from Scale-Invariant Keypoints", International Journal of Computer Vision, vol. 60, no.2, pp. 91-110, 2004.

[2] Bay, Herbert, Tinne Tuytelaars, and Luc Van Gool. "Surf: Speeded up robust features", in European Conference on Computer Vision (ECCV). Springer, Berlin, Heidelberg, 2006.

[3] E. Rublee, V. Rabaud, K. Konolige and G. Bradski, "Orb: An efficient alternative to sift or surf", in 2011 International Conference on Computer Vision, 2011, pp. 2564–2571.

[4] K. Mikolajczyk and C. Schmid, "A performance evaluation of local descriptors," IEEE Transactions on Pattern Analysis and Machine Intelligence, vol.27, no.10, pp.1615-1630, 2005.

[5] V. Balntas, L. Tang and K. Mikolajczyk, "Bold-binary on-line learned descriptor for efficient image matching", in Computer Vision and Pattern Recognition (CVPR), pp. 2367-2375, 2015.

[6] Luca Cavalli, Viktor Larsson, Martin Ralf Oswald, Torsten Sattler, and Marc Pollefeys, "Handcrafted Outlier Detection Revisited", in European conference on computer vision, 2020.

[7] Kwang Moo Yi, Eduard Trulls, Yuki Ono, Vincent Lepetit, Mathieu Salzmann, and Pascal Fua, "Learning to find good correspondences", in Computer Vision and Pattern Recognition (CVPR), 2018.

[8] Zheng Dang, Kwang Moo Yi, Yinlin Hu, Fei Wang, Pascal Fua and Mathieu Salzmann, "Training without eigen decomposition of deep networks with zero eigenvalue-based losses", European Conference on Computer Vision (ECCV), 2018.

[9] Rene Ranftl and Vladlen Koltun, "Deep fundamental matrix estimation", in European Conference on Computer Vision (ECCV), 2018.

[10] Eric Brachmann and Carsten Rother, "Neural-guided ransac: Learning where to sample the model hypotheses", in International Conference on Computer Vision (ICCV), 2019.

[11] Edward Johns and Guang-Zhong Yang, "Ransac with 2d geometric cliques for image retrieval and place recognition", in Computer Vision and Pattern Recognition Workshops (CVPRW), 2015.

[12] Daniel Barath and Jir Matas, "Graph-cut ransac", in Computer Vision and Pattern Recognition (CVPR), 2018.

[13] Daniel Barath, Jiri Matas, and Jana Noskova, "Magsac: marginalizing sample consensus", in Computer Vision and Pattern Recognition (CVPR), 2019.

[14] Martin A Fischler and Robert C Bolles, "Random sample consensus: a paradigm for model fitting with applications to image analysis and automated cartography", Communications of the ACM, vol. 24, pp. 381–395, 1981.

[15] Xiaomeng Wu and Kunio Kashino, "Adaptive dither voting for robust spatial verification", in International Conference on Computer Vision (ICCV), 2015.

[16] Xiaomeng Wu and Kunio Kashino, "Robust spatial matching as an ensemble of weak geometric relations", in British Machine Vision Conference (BMVC), 2015.

[17] Johannes L Schonberger, True Price, Torsten Sattler, Jan-Michael Frahm, and Marc Pollefeys, "A vote-and-verify strategy for fast spatial verification in image retrieval", in Asian Conference on Computer Vision (ACCV), 2016.

[18] Jui-Man Chiu, Zen Chen, Jen-Hui Chuang, Tsorng-Lin Chia, "Determination of feature correspondences in stereo images using a calibration polygon", Pattern Recognition, vol. 30, pp. 1387-1400, 1997.

[19] Dubrofsky, Elan, "Homography estimation", Diplomová práce. Vancouver: Univerzita Britské Kolumbie, vol. 5, 2009.

[20] J. Demmel, I. Dumitriu, and O. Holtz, "Fast linear algebra is stable", Numer. Math, pp. 59–91, 2007.

[21] Amila Jakubovic and Jasmin Velagic, "Image feature matching and object detection using brute-force matchers", in International Symposium ELMAR, pp. 83–86, 2018.